\documentclass[letterpaper]{article} 
\usepackage{aaai25}  
\usepackage{times}  
\usepackage{helvet}  
\usepackage{courier}  
\usepackage[hyphens]{url}  
\usepackage{graphicx} 
\usepackage{natbib}  
\usepackage{caption} 
\usepackage{algorithm}
\usepackage{algorithmic}
\newtheorem{definition}{Definition}
\newtheorem{task}{Task}
\usepackage{booktabs}
\usepackage{tabularx}

\usepackage{newfloat}
\usepackage{listings}
\DeclareCaptionStyle{ruled}{labelfont=normalfont,labelsep=colon,strut=off} 
\floatstyle{ruled}
\newfloat{listing}{tb}{lst}{}
\floatname{listing}{Listing}
\title{Exploring Iterative Enhancement for Improving Learnersourced Multiple-Choice Question Explanations with Large Language Models}
\author{
    Qiming Bao\textsuperscript{\rm 1, 2}, Juho Leinonen\textsuperscript{\rm 3}, Alex Yuxuan Peng\textsuperscript{\rm 1}, Wanjun Zhong\textsuperscript{\rm 4}, Gaël Gendron\textsuperscript{\rm 1}, Tim Pistotti\textsuperscript{\rm 1}, Alice Huang\textsuperscript{\rm 5}, Paul Denny\textsuperscript{\rm 3}, Michael Witbrock\textsuperscript{\rm 1}, Jiamou Liu\textsuperscript{\rm 1}
    \thanks{With help from the AAAI Publications Committee.}
}
\affiliations{
    \textsuperscript{\rm 1}Strong AI Lab, NAOInstitute, Waipapa Taumata Rau - The University of Auckland\\
    \textsuperscript{\rm 2}Xtracta, New Zealand\\
    \textsuperscript{\rm 3}School of Computer Science, University of Auckland\\
    \textsuperscript{\rm 4}School of Computer Science and Engineering, Sun Yat-Sen University\\
    \textsuperscript{\rm 5}School of Life and Environmental Sciences, University of Sydney\\
    \{qbao775,ypen260,ggen187\}@aucklanduni.ac.nz, \{juho.leinonen,p.denny,m.witbrock,jiamou.liu\}@auckland.ac.nz, zhongwj25@mail2.sysu.edu.cn, alice.huang@sydney.edu.au
}

\usepackage{bibentry}

\begin{document}

\maketitle

\begin{abstract}
Large language models (LLMs) have demonstrated strong capabilities in language understanding and generation, and their potential in educational contexts is increasingly being explored.
One promising area is learnersourcing, where students engage in creating their own educational content, such as multiple-choice questions.
 A critical step in this process is generating effective explanations for the solutions to these questions, as such explanations aid in peer understanding and promote deeper conceptual learning.   
However, students often find it difficult to craft high-quality explanations due to limited understanding or gaps in their subject knowledge. 
To support this task, we introduce ``ILearner-LLM,'' a framework that uses iterative enhancement with LLMs to improve generated explanations. 
The framework combines an explanation generation model and an explanation evaluation model fine-tuned using student preferences for quality, where feedback from the evaluation model is fed back into the generation model to refine the output. 
Our experiments with LLaMA2-13B and GPT-4 using five large datasets from the PeerWise MCQ platform show that ILearner-LLM produces explanations of higher quality that closely align with those written by students. Our findings represent a promising approach for enriching the learnersourcing experience for students and for leveraging the capabilities of large language models for educational applications.
\end{abstract}

%

\section{Introduction}

Given the remarkable performance of large language models (LLMs) in understanding and generating natural language~\citep{wei2022emergent,brown2020language}, they appear to offer great potential across many applications within education. Learnersourcing, a pedagogical approach that distributes the task of generating learning content among students, leverages their collective intelligence to enhance the learning experience~\citep{jiang2018review,KHOSRAVI2023100151,kim2015learnersourcing}. On platforms such as PeerWise~\citep{denny2008peerwise} and RiPPLE~\citep{khosravi2019ripple}, learnersourcing often involves students creating multiple-choice questions and providing corresponding explanations. 
However, crafting high-quality explanations requires a deep understanding of the underlying concepts, a challenge that students may struggle with. Additionally, because it is often not mandatory for students to include explanations when generating questions, they may choose to omit them altogether.
The automatic generation and evaluation of high-quality explanations can serve as a scaffold for learners, offering tailored support that fosters deeper understanding and greater independence, particularly in the context of learnersourcing tasks where students collaboratively create and refine educational content.


The main challenges in automatic explanation generation within this context are driven by several key factors.  
First, one hurdle is simulating the way students write explanations and generating text that aligns with what students value in a well-crafted explanation.
This challenge goes beyond replicating content; it requires capturing the characteristics of how students express their understanding.
Second, the scarcity of high-quality datasets that include explanations poses another major challenge. Since writing explanations is often not mandatory for students in learnersourcing platforms, there is a limited amount of annotated data available for training models. This lack of data makes it difficult to achieve high performance in automatic explanation generation. 


In this work we aim to use LLMs to auto-generate explanations for student questions.
The integration of automatic explanation generation using LLMs in learnersourcing may offer multiple advantages. Firstly, instant feedback from large language models has the potential to boost students' learning efficiency~\citep{dai2023can}. Secondly, interaction with such models and observing their outputs can help to promote learner autonomy~\citep{yildiz2023conversational}. Thirdly, the use of pretrained large-scale models facilitates the generation of multi-faceted and comprehensive learnersourced content~\citep{khosravi2023learnersourcing}. Additionally, the LLMs can be fine-tuned to student preferences by using data on how students rate questions~\citep{ni2022deepqr}. Despite these benefits, employing LLMs in a learnersourcing context also presents challenges.  In particular, there is limited access to high-quality student-written explanations which makes it difficult to fine-tune models to generate explanations that are both linguistically and semantically similar to those written by students~\citep{KHOSRAVI2023100151}. 

We present and evaluate a framework, ILearner-LLM, that generates explanations in an iterative fashion. ILearner-LLM makes use of two LLMs fine-tuned on data sourced from PeerWise, a popular learnersourcing platform.  We use a \emph{generation model} that generates an explanation from a given question, and an \emph{evaluation model} that rates the quality of an explanation to a question. At each iteration, ILearner-LLM applies instruction prompting to generate an explanation, and then evaluates the quality of the generated explanation by outputting a quality rating score. The quality rating is injected into the instruction prompt for the explanation generation model in the next iteration. This process is repeated multiple times in order to iteratively generate higher-quality explanations that are linguistically and semantically similar to those written by students. We summarise our main findings as follows:

\begin{itemize}
    \item Our iterative enhancement framework, ``ILearner-LLM'', implemented with LLaMA2-13B and GPT-4, demonstrates notable improvements over using the models without the framework in generating explanations that closely resemble those  written by students, as evidenced by BLEU and BERT scores.
    
    \item We find that ILearner-LLM can help instruction fine-tuned LLaMA2-13B achieve greater improvement 
    compared to applying ILearner-LLM to GPT-4. 
    This is mainly because LLaMA2-13B, which was fine-tuned using the same instruction set employed in iterative prompting, better aligns with the instructional framework, enabling the model to learn and perform more effectively through multiple iterations.   
We also evaluated the impact of feeding both the generated explanation and the quality rating score from either the most recent iteration or all iterations into GPT-4. Our results show no significant difference between these approaches.
These findings suggest that ILearner-LLM, particularly when combined with instruction fine-tuning, is more effective in generating student-like explanations compared to existing models.

\item We find that the evaluation models which have been instruction fine-tuned on explanations rated by students demonstrate lower MSE scores compared to models that have not been fine-tuned. The model that has been fine-tuned on a more diverse range of subjects and additional data (Merged) achieves even lower MSE scores. 
The (Merged) training set indicates that the training sets from all subjects have been merged together.


\end{itemize}


\section{Related Work}

Artificial General Intelligence (AGI) aims to enable machines to understand, learn, and apply knowledge as broadly as humans do~\citep{goertzel2014artificial}. Making machines think and act in a more human-like manner is crucial for the development of AGI ~\citep{goertzel2012architecture,lake2017building,ouyang2022training}. Prompting is a key method for enabling complex human-AI interactions, bridging the gap from AGI concepts to practical uses by fostering adaptability and understanding in line with AGI goals~\citep{hao-etal-2023-reasoning,madaan2024self}. Chain-of-Thought (CoT) prompting~\citep{wei2022chain} has been introduced to not only generate answers but also the intermediate steps. ITER-RETGEN~\citep{shao-etal-2023-enhancing} integrates CoT prompting with a retriever to iteratively decompose complex queries, enhancing performance on multi-hop question answering tasks. The LLM-Augmenter~\citep{peng2023check} uses an agent to interact with external knowledge, helping large language models improve open-domain question answering and reduce hallucinations. Iterative prompting has been developed to refine translation results from large language models and decrease ``translationese''~\citep{chen2023iterative}.

In the domain of automated explanation generation and question quality evaluation using deep learning, research remains sparse. Building a natural language generation system with explanations is an ongoing goal~\cite{reiter1997building}. Template or rule-based methods~\cite{holmes1994text}, knowledge base methods~\cite{wang2010got}, and Long Short-Term Memory (LSTM) networks~\cite{costa2018automatic} have been used for automatic explanation generation to build explainable recommender systems. With the development of pre-trained transformer-based language models and stronger representation learning for understanding context, BERT-based models have been employed to assess the convincingness of learner-generated explanations, a key criterion for quality~\citep{bhatnagar2020learnersourcing}. However, this approach is designed to evaluate the convincingness of explanations for peers, which requires humans to label data. It does not directly evaluate a single explanation. Recently, a transformer model has been trained with contrastive learning to evaluate question quality, incorporating various elements such as question context and distractors~\citep{ni2022deepqr}. Despite its merits, this method necessitates manual feature engineering, which includes explicity-defined features as the model input such as readability, clarity, length of distractors and number of distractors. Contrary to the previous work, explanation evaluation in ILearner-LLM is performed using LLMs fine-tuned to predict the quality rating given an explanation and a question, without relying on explicitly defined features.  We also did not rely on a self-reward model as described in~\citep{yuan2024self}, nor did we apply the rationale generation method outlined in~\citep{hsieh-etal-2023-distilling}. Additionally, our goal of incorporating explanation evaluation is to iteratively improve the quality of generated explanations by feeding the quality ratings back into the explanation generation model.

\section{Problem Formulation}\label{sec:problem}
In this section, we formally define the {\em multiple-choice question explanation generation} and {\em evaluation} tasks. 
When authoring an MCQ in a learnersourcing system like PeerWise~\citep{denny2008peerwise}, a student needs to specify seven components: a question stem, a correct answer, (up to) four distractors, and a paragraph that explains the idea and rationale behind the question. 
The question is then submitted to an online repository of MCQs accessible by the class. After answering a question, a student may leave a holistic quality rating (on a 6-point scale from $0,1,\ldots,5$) by considering the  
``\textit{language, quality of distractors, quality of explanation, and relevance to the course}'' as suggested by the PeerWise platform~\citep{denny2008peerwise}.


\begin{definition}(Multiple-Choice Questions (MCQs)) MCQs are a set of questions, $\{M_1,M_2,\ldots,M_n\}$, collected from a course, where each $M_i$ consists of a stem $S_i$, a correct answer $A_i$, distractors $D_{i,j}$ where $j\in \{1,2,3,4\}$, explanation $E_i$, and is assigned a rating $r_i$. 
\end{definition}

\begin{task}(MCQ explanation generation) Multiple-Choice Question (MCQ) explanation generation aims to construct a model, $\mathsf{G}$, which takes the question stem $S_i$, the correct answer $A_i$, and distractors $D_{i,j}$ as inputs, and produces a generated explanation $E_i$ as the output.
\end{task}

\begin{task}(MCQ explanation evaluation) The goal of Multiple-Choice Question (MCQ) explanation evaluation is to build a model, $\mathsf{G}$, which takes as input the question stem $S_i$, the correct answer $A_i$, distractors $D_{i,j}$, and the generated explanation $E_i$, and outputs a quality rating $r_i$ for the MCQ. 
\end{task}

\section{Method}\label{sec:methods}
\subsection{Iterative MCQ Explanation Enhancement}
\begin{figure*}[ht]
\centering
\includegraphics[width=\textwidth]{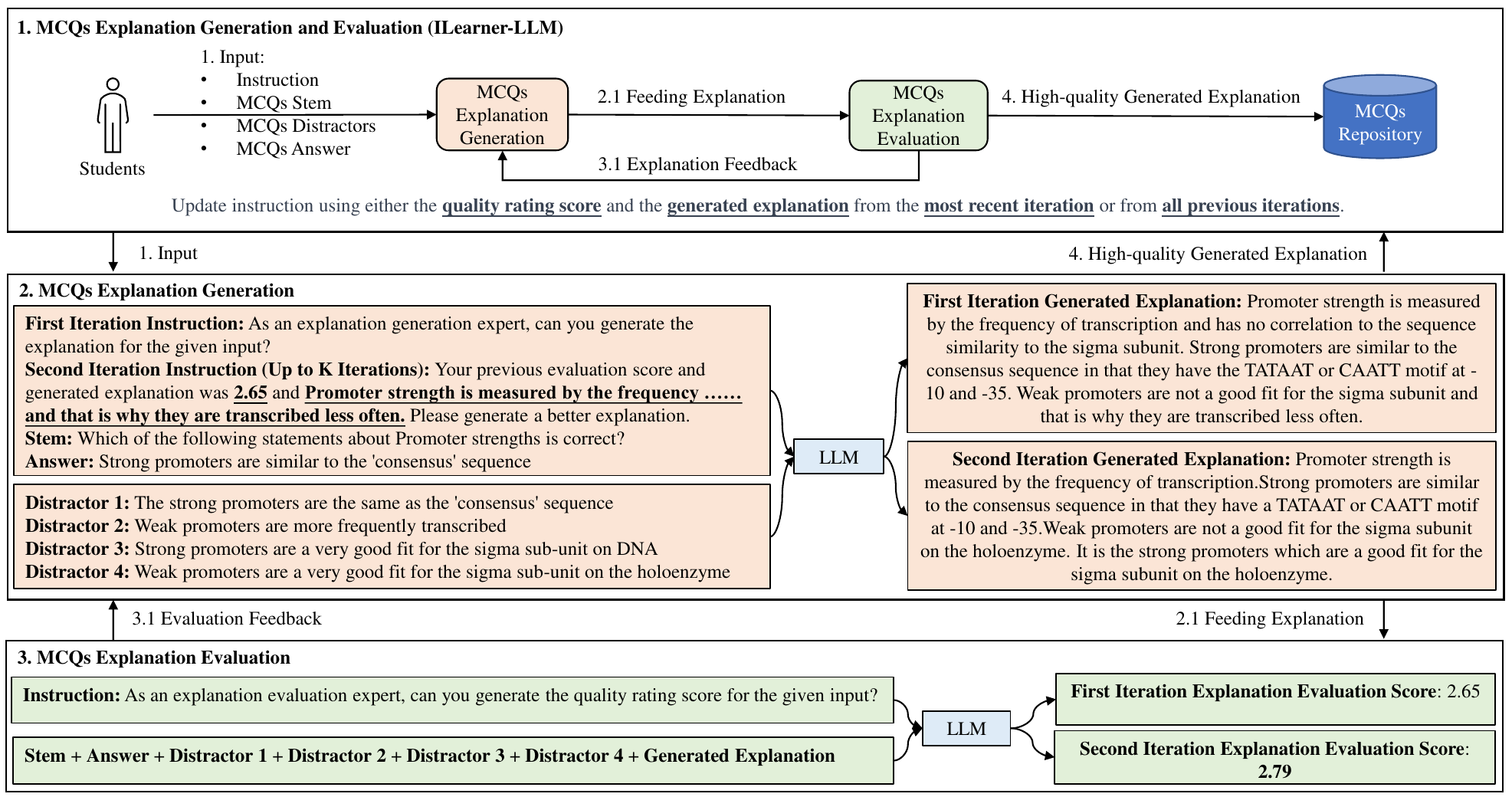}
\caption{Architecture of the iterative enhancement framework ``ILearner-LLM'' using large language models for multiple-choice question explanation generation and evaluation.}
\label{system-architecture}
\end{figure*}

The architecture of our system is illustrated in Figure~\ref{system-architecture}. The \textbf{MCQ Explanation Generation Module} is implemented through instruction fine-tuning to automatically generate explanations for MCQs. The generated explanations are then provided as inputs to the \textbf{MCQ Explanation Evaluation Module}. This module is implemented through instruction fine-tuning, enabling it to automatically assess the quality of the generated explanations. The  generation module and evaluation module will interact for up to K iterations, where K is a hyper-parameter. The evaluation score from the MCQ Explanation Evaluation Module and the generated explanation from the most recent iteration will be fed back to the MCQ Explanation Generation Module, which in turn prompts the model to generate a new explanation. This iterative process continues until it reaches the predefined number of iterations, K. Our ILearner-LLM framework allows the inclusion of either just the generated explanation and rating score from the most recent iteration, or the generated explanation and rating score from all previous iterations, the latter of which is more appropriate for models that can support long sequence input. 
The pseudocode for our ILearner-LLM MCQ Explanation Generation and Evaluation framework is shown in Algorithm~\ref{MCQs-explanation-generation-evaluation-code}. 


\begin{algorithm}[h]
	\caption{MCQ Explanation Generation and Evaluation} 
	\label{MCQs-explanation-generation-evaluation-code} 
	\begin{algorithmic}
		\REQUIRE pre-defined iteration step $K$, initial iteration\_step = 0, multiple-choice questions (MCQs) $\{M_1,M_2,\ldots,M_n\}$, question stem $S_i$, a correct answer $A_i$, distractors $D_{i,j}$ where $j\in \{1,2,3,4\}$, explanation $E_i$, large language model (LLM), batch\_size bs, learning\_rate lr, history = []\\
		\emph{\textbf{1. MCQ explanation generation instruction fine-tuning}} \\
        \FOR{instruction, $S_i$, $A_i$, $D_{i,j}$ from MCQs}{
        \item LLM, Loss = \textbf{next\_token\_prediction}(LLM, instruction, $S_i$, $A_i$, $D_{i,j}$)}\ENDFOR \\
        \emph{\textbf{2. MCQ explanation evaluation instruction fine-tuning}} \\
        \FOR{instruction, $S_i$, $A_i$, $D_{i,j}$, $E_i$ from MCQs}{
        \item LLM, Loss = \textbf{next\_token\_prediction}(LLM, instruction, $S_i$, $A_i$, $D_{i,j}$, $E_i$)}\ENDFOR \\
        \emph{\textbf{3. Iterative MCQ explanation enhancement}} \\
        \WHILE{iteration\_step $<$ $K$}{
        \item reg\_explanation = \textbf{explanation\_generator}(instruction, $S_i$, $A_i$, $D_{i,j}$)} \\
        \item rating\_score = \textbf{explanation\_evaluator}(instruction, $S_i$, $A_i$, $D_{i,j}$, $E_i$) \\
        \IF{only use generated explanation and rating score from the \textbf{most recent iteration}}
         \STATE instruction += reg\_explanation + rating\_score + ``Please generate a better explanation.''
         \ELSIF{use the generated explanation and rating score from the \textbf{all previous iterations}} 
         \STATE history.append(reg\_explanation,rating\_score)\\
         \STATE instruction = instruction.join(history) + ``Please generate a better explanation.''
        \ENDIF
        \item iteration\_step = iteration\_step + 1\ENDWHILE \\
	\end{algorithmic} 
\end{algorithm}

\subsection{MCQ Explanation Generation}
As depicted in Figure~\ref{system-architecture}, we conduct instruction fine-tuning to train a model for generating explanations for MCQs, and then use instruction prompting with the well-trained model to generate these MCQ explanations. Instruction Fine-Tuning and Instruction Prompting adapt a pretrained model to follow specific input instructions more accurately. The difference lies in the fact that instruction fine-tuning involves additional training with examples that pair these instructions with desired outputs, thereby enhancing the model's task-specific performance~\citep{mishra2021cross,wei2021finetuned}. In contrast, instruction prompting does not explicitly train the model; instead, it uses the instructions as part of the prompt for a pre-trained model. The instructions utilised for generating explanations and conducting evaluations are delineated in the system architecture, as depicted in Figure~\ref{system-architecture}. The model inputs include the instruction, question stem, correct answer, and distractors. The model outputs the explanation for the MCQ. During data preprocessing of the five PeerWise datasets
(Sydney Biology Subject, Cardiff Biology Subject, Auckland Law Subject, UK Medical Year 1 Subject, and UK Medical Year 2 Subject), we retained only  MCQs with quality rating scores of 3 or higher and explanations that are longer than 10 words. This step is undertaken to provide some simple quality control for the MCQs in the training set.

The instruction of the initial iteration is formalised as ``As an explanation generation expert, can you generate an explanation for the given input?''. The further iteration instruction (up to K iterations) is formalised as ``Your previous evaluation score and generation explanation was \underline{the most recent iteration explanation score} and \underline{the most recent iteration generated explanation}. Please generate a better explanation''. 

\subsection{MCQ Explanation Evaluation}
Similar to the module above, we employ instruction fine-tuning to train a large language model to evaluate the generated explanations. In the absence of quality rating scores for explanations, we trained an evaluation model for MCQ explanations using the quality rating scores derived from merging five PeerWise MCQ training sets. The model's input comprises the instruction, question stem, correct answer, distractors, and the explanation. The model's output is the quality rating score for the MCQs. The instruction used in the MCQ Explanation Evaluation Module is, ``As an explanation evaluation expert, can you generate the quality rating score for the given input?''.

Whenever the MCQ Explanation Evaluation Module predicts a quality rating score, the MCQ Explanation Generation Module is prompted to regenerate the explanation. This regenerated explanation then replaces the one from the previous iteration. The new explanation, along with other inputs, is subsequently fed back into the MCQ Explanation Evaluation Model for re-evaluation. This cycle continues until the number of iteration steps surpasses the predefined K.

\section{Experiments}\label{sec:experiment}
\subsection{Experiment Setup}
\begin{table*}[h]
\centering
\resizebox{\textwidth}{!}{
\begin{tabular}{@{}lcccc@{}}
\toprule
Models               & \# Iteration Step & Avg Quality Rating Score & Avg BLEU Score  & Avg BERT Score  \\ \midrule
\multicolumn{5}{c}{Sydney Biology Subject}                                                                     \\
LLaMA2-13B Merged        & 1                 & 2.84                            & 34.34          & 61.62          \\
LLaMA2-13B Merged ILearner-LLM & \textbf{2.37}              & 2.87                   & \textbf{38.07} & 62.00 \\
GPT-4              & 1                 & 3.02                            & 34.24          & \textbf{63.72}          \\
GPT-4 ILearner-LLM      & 1.63              & 3.12                   & 35.19 & 63.45 \\
GPT-4 ILearner-LLM All History     & 1.70              & \textbf{3.14}                   & 35.08 & 63.58 \\
\midrule
\multicolumn{5}{c}{Cardiff Biology Subject}                                                                    \\
LLaMA2-13B Merged        & 1                 & 3.07                            & 25.59          & 58.60          \\
LLaMA2-13B Merged ILearner-LLM & \textbf{2.08}              & 3.11                   & \textbf{30.58} & 58.27 \\
GPT-4              & 1                 & 3.18                            & 29.08          & 58.72          \\
GPT-4 ILearner-LLM      & 1.84              & \textbf{3.23}                   & 29.91 & 58.57 \\ 
GPT-4 ILearner-LLM All History     &    1.36           &    3.21                & 30.43 & \textbf{58.77} \\
\midrule
\multicolumn{5}{c}{Auckland Law Subject}                                                                       \\
LLaMA2-13B Merged        & 1                 & 4.11                            & 27.82          & 58.01          \\
LLaMA2-13B Merged ILearner-LLM & \textbf{2.23}              & 4.20                   & \textbf{34.33} & \textbf{59.95} \\
GPT-4              & 1                 & 4.22                            & 24.31          & 57.19          \\
GPT-4 ILearner-LLM      & 1.74              & \textbf{4.29}                   & 24.09 & 56.91 \\ 
GPT-4 ILearner-LLM All History     &   1.45            &    \textbf{4.29}                & 24.26 & 57.11 \\
\midrule
\multicolumn{5}{c}{UK Medical Year 1 Subject}                                                                  \\
LLaMA2-13B Merged        & 1                 & 3.07                            & 27.60          & 58.45          \\
LLaMA2-13B Merged ILearner-LLM & \textbf{2.18}              & 3.09                   & \textbf{32.52} & 59.06 \\
GPT-4              & 1                 & 3.20                            & 28.29          & \textbf{59.47}          \\
GPT-4 ILearner-LLM      & 1.60              & \textbf{3.23}                   & 28.65 & 59.38 \\ 
GPT-4 ILearner-LLM All History     &   1.27            & 3.21                   & 29.10 & 59.43 \\
\midrule
\multicolumn{5}{c}{UK Medical Year 2 Subject}                                                                  \\
LLaMA2-13B Merged        & 1                 & 3.05                            & 23.89          & 56.82          \\
LLaMA2-13B Merged ILearner-LLM & \textbf{2.44}              & 3.06                   & 30.43 & 56.96 \\
GPT-4              & 1                 & 3.15                            & 30.67          & 58.17          \\
GPT-4 ILearner-LLM      & 1.88              & \textbf{3.18}                   & 31.63 & 57.97 \\ 
GPT-4 ILearner-LLM All History     &    1.53           &    \textbf{3.18}                & \textbf{31.83} & \textbf{58.21} \\
\bottomrule
\end{tabular}}
\caption{In an experiment, we evaluated two models, fine-tuned LLaMA2-13B (Merged) and GPT-4, for generating MCQ explanations. The evaluation used the fine-tuned LLaMA2-13B (Merged) model. The ``ILearner-LLM All History'' model incorporates explanations and scores from all previous iterations, whereas the ILearner-LLM framework model uses only the most recent explanation and score.}
\label{explanation-generator-evaluator-comparison}
\end{table*}

\paragraph{Datasets} A typical multiple-choice question (MCQ) on PeerWise, a free learnersourcing platform employed by more than 2,500 universities worldwide~\citep{denny2008peerwise}, includes the following components: a question stem, an answer, distractors, and an explanation. For each question, there is only one correct answer.  Each question also has an average quality score, as rated by students, in the range from 0 to 5. The explanation, provided by the student who created the question, ideally demonstrates the background knowledge needed and the steps involved in solving the question. 

We conducted our experiment on five learnersourced multiple-choice question datasets, covering three academic subjects:  biology, law, and medicine, exported from the PeerWise platform. We selected these datasets because they contain a large number of questions. To improve reliability, only questions that received at least 10 ratings were included. The average explanation length corresponds to the number of words per sentence.


\begin{table}[h]
\centering
\Large
\resizebox{\columnwidth}{!}{
\begin{tabular}{@{}lccc@{}}
\toprule
Subject         & Sydney Biology       & Cardiff Biology      & Auckland Law     \\ \midrule
\# MCQs          &    2311                  & 6955                     &     3449                 \\
\# Ratings      &   57585                   & 581937                     &   65645                   \\
\# Ratings/MCQ &   24.91                   &   83.67                   &  19.03                    \\
Avg exp length &  108.82                    &   75.09                   &   48.13                   \\ \midrule
Subject         & UK Medical Year 1         & \multicolumn{2}{c}{UK Medical Year 2}     \\ \midrule
\# MCQs          & \multicolumn{1}{c}{3991} & \multicolumn{2}{c}{2789}  \\
\# Ratings      & \multicolumn{1}{c}{305067} & \multicolumn{2}{c}{271524} \\
\# Ratings/MCQ & \multicolumn{1}{c}{76.43} & \multicolumn{2}{c}{97.35} \\
Avg exp length & \multicolumn{1}{c}{68.94} & \multicolumn{2}{c}{83.38} \\ \bottomrule
\end{tabular}}
\caption{Details on the PeerWise datasets used for conducting explanation generation experiment.}
\label{dataset_details}
\end{table}

\begin{table*}[h]
\centering
\begin{tabular}{@{}lcccccc@{}}
\toprule
\begin{tabular}[c]{@{}l@{}}Models $\rightarrow$ \\ Metrics $\downarrow$\end{tabular} & Vicuna-13B & \begin{tabular}[c]{@{}c@{}}Fine-tuned\\ Vicuna-13B\end{tabular} &\begin{tabular}[c]{@{}c@{}}Fine-tuned\\ LLaMA2-13B\end{tabular} &\begin{tabular}[c]{@{}c@{}}Fine-tuned\\ LLaMA2-13B\\Merged\end{tabular} & GPT-3.5 & GPT-4  \\ \midrule
\multicolumn{7}{c}{Sydney Biology Subject}                                                                                                        \\
Avg BLEU Score                                            & 8.59   &  33.91 & \textbf{34.80} & 34.34&   30.25                                                    & 34.24  \\
Avg BERT Score                                            & 20.17     & 63.33 & 62.26  & 61.62 &  63.56                                                    & \textbf{63.72}  \\ \midrule
\multicolumn{7}{c}{Cardiff Biology Subject}                                                                                                       \\
Avg BLEU Score                                            &   3.36  &   15.33 & 25.37 & 25.59 &  25.65                                                   & \textbf{29.08}  \\
Avg BERT Score                                            &  8.76    &  51.72  & 56.85 & 58.60 &  57.69                                                   & \textbf{58.72}  \\ \midrule
\multicolumn{7}{c}{Auckland Law Subject}                                                                                                          \\
Avg BLEU Score                                            &  3.09    &  9.36  & 26.39 & \textbf{27.82} &  22.16                                                  & 24.31  \\
Avg BERT Score                                            &  7.99    &  45.38 & 57.07 & \textbf{58.01} &  57.11                                                     & 57.19  \\ \midrule
\multicolumn{7}{c}{UK Medical Year 1 Subject}                                                                                                      \\
Avg BLEU Score                                            &  1.92    & 15.09 & 26.17 & 27.60 &   25.44                                                    & \textbf{28.29}  \\
Avg BERT Score                                            &  6.22    & 52.06 & 57.23 & 58.45 & 58.44                                                      & \textbf{59.47}  \\ \midrule
\multicolumn{7}{c}{UK Medical Year 2 Subject}                                                                                                   \\
Avg BLEU Score                                            &  4.23    &   17.72 & 24.76  &  23.89 & 26.61                                                      & \textbf{30.67}  \\
Avg BERT Score                                            &  12.47    &   51.62 & 55.91 & 56.82 &   57.15                                                     & \textbf{58.17}  \\ \bottomrule
\end{tabular}
\caption{We compared the performance of fine-tuned and non-fine-tuned Vicuna-13B, fine-tuned LLaMA2-13B, and GPT-4 on 100 MCQ explanation test cases from Biology, Law, and Medical subjects in Sydney, Cardiff, Auckland, and the UK.}
\label{explanation-generator-comparison}
\end{table*}

\begin{table}[h]
\centering
\resizebox{\columnwidth}{!}{
\begin{tabularx}{\columnwidth}{@{}l *{6}{X}@{}}
\toprule
Iteration Steps $\rightarrow$   & 1   & 2  & 3 & 4 & 5 & 6 \\
Models $\downarrow$            &     &    &   &   &   &   \\ \midrule
\multicolumn{7}{c}{Sydney Biology Subject}    \\
LLaMA2-13B Merged ILearner-LLM         &  38   &  26  & 14  &  11 & 5  & 6  \\ 
GPT-4 ILearner-LLM             &  61   &  29  & 3  &  2 &  3 & 2  \\
GPT-4 ILearner-LLM All History            &  50   &  40  & 4  &  3 &  2 & 1  \\
\midrule
\multicolumn{7}{c}{Cardiff Biology Subject}   \\
LLaMA2-13B Merged ILearner-LLM         &  36   & 38   & 15  & 5  & 5  & 1  \\ 
GPT-4 ILearner-LLM             &  63   & 17   &  8 &  3 &  3 & 6  \\
GPT-4 ILearner-LLM All History            &  75   &  20  & 1  & 3  & 0  & 1 \\\midrule
\multicolumn{7}{c}{Auckland Law Subject}      \\
LLaMA2-13B Merged ILearner-LLM         &  27   & 44   &  18 &  4 & 4  & 3  \\ 
GPT-4 ILearner-LLM             &  65   & 18   & 4  & 6  & 5  & 2  \\
GPT-4 ILearner-LLM All History            & 72    &  20  & 4  & 1  & 1  & 2  \\\midrule
\multicolumn{7}{c}{UK Medical Year 1 Subject} \\
LLaMA2-13B Merged ILearner-LLM         &  37   & 35  & 12  &  8 &  5 &  3 \\ 
GPT-4 ILearner-LLM             &  74   &  10  &  7 &  4 & 1  & 4  \\
GPT-4 ILearner-LLM All History     &  81   &  12  & 6  & 1  & 0  & 0           \\\midrule
\multicolumn{7}{c}{UK Medical Year 2 Subject} \\
LLaMA2-13B Merged ILearner-LLM         &  28   &  35  & 15  & 12  &  7 &  3 \\ 
GPT-4 ILearner-LLM             &  58   &  22  & 9  &  2 & 3  & 6  \\
GPT-4 ILearner-LLM All History            &  65   & 24   & 8  & 0  & 2  &  1 \\\bottomrule
\end{tabularx}}
\caption{Comparative analysis of iterative enhancement framework performance: number of iterations required for optimal quality rating score, BLEU, and BERT Scores against student-written ground truth.}
\label{iteration-analysis}
\end{table}

\paragraph{Models} We select the large language models LLaMA2-13B~\citep{touvron2023llama} and GPT-4~\citep{openai2023gpt4} as the backbone models for conducting the main experiments, which include instruction fine-tuning and prompting for the generation and evaluation of multiple-choice question (MCQ) explanations. We chose Vicuna-13B~\citep{vicuna2023} and GPT-3.5~\citep{openai2023chatgpt} as baseline models.

\paragraph{Data Preprocessing} For the MCQ Explanation Generation module and the MCQ Explanation Evaluation module, we employ different data preprocessing strategies. To train the Explanation Generator, which aids in generating high-quality explanations, resulting in a total of 19,495 questions after filtering out those quality rating below 3, a length of less than 10, the inclusion of an image in the question stem and fewer than 10 ratings. See Table~\ref{dataset_details} for details on the  datasets. We restrict the length of the explanations because we find that many short explanations are incomplete. 
We trained the explanation evaluator using 27,140 high-rated and low-rated questions across all subjects. 

\paragraph{Settings} We conducted all the instruction fine-tuning for Vicuna-13B and LLaMA2-13B MCQ explanation generation and evaluation experiments on 8 NVIDIA A100 GPUs with 80G GPU memory. We trained our model for 5 epochs, using a batch size of 1 and a maximum sequence length of 512. We set the learning rate to 2e-05 and the warmup ratio to 0.03. To leverage the power of multi-GPUs, we utilised the torchrun tool for training. The source
code is available \footnote{\url{https://github.com/Strong-AI-Lab/Explanation-Generation}}.

\subsection{Iterative MCQ Explanation Enhancement} We iterate the process of explanation generation and evaluation over K steps, with each iteration comprising one instance of explanation generation and evaluation. We recorded the score for each evaluation, and the similarity between the generated explanation and the original explanation written by the student. We then computed the number of iterations required to improve the evaluation score, the generated explanation, and the similarity to the original student-written explanation. In our experiment, we set a number of iteration steps, K=5, for halting iterations: the model generates explanations iteratively over K iterations. In each iteration, the model feeds the previously generated explanation and the quality rating score into the instruction and prompts the MCQ explanation generation module. The specific results are shown in Table~\ref{explanation-generator-evaluator-comparison}. 

For each generation, we calculate the normalized average score for the question quality rating, BLEU score, and BERT score across the iteratively generated K explanations. We then select the explanation with the highest normalized average score. The question quality rating score is normalized from a float value to a range between 0 and 1.  Our goal is to determine how many iterations are required to surpass the explanation generated in the first iteration.  
Using our iterative enhancement framework, ILearner-LLM, we found that approximately 2.26 iterations are required for the fine-tuned LLaMA2-13B model, and 1.73 iterations for GPT-4, when generating explanations using only the most recent generated explanation and its quality rating score.  For GPT-4, when including all previously generated content along with their quality rating scores, an average of 1.46 iterations are required to produce an explanation that surpasses the original in terms of question quality rating, BLEU score, and BERT score.


ILearner-LLM, which incorporates LLaMA2-13B and GPT-4 models, shows notable improvements in generating higher quality explanations that are both syntactically and semantically closer to those written by students across the five PeerWise datasets. It outperforms a fine-tuned LLaMA2-13B by 5.33 in BLEU score and surpasses GPT-4 by 3.86. Furthermore, since GPT-4 only supports up to 8K tokens in the input, we applied ILearner-LLM iteratively by feeding the generated explanations and quality rating scores in separate inputs from all the previous history into GPT-4, which achieved a 0.82 and 0.05 improvement over GPT-4 on BLEU and BERT score. These results demonstrate that ILearner-LLM, with instruction fine-tuning, is more effective in generating higher-quality explanations that are syntactically and semantically closer to the explanations written by students compared to existing models.

We conducted an analysis of our proposed iterative enhancement framework using different numbers of iterations as shown in Table~\ref{iteration-analysis}. We recorded the number of iterations required to find an explanation with the highest rating score, as well as the highest BLEU and BERT scores, compared to the ground truth of student-written explanations. ILearner-LLM can assist the fine-tuned LLaMA2-13B Merged in generating better explanations over a greater number of iteration steps compared to GPT-4. GPT-4 produces high-quality explanations without task-specific fine-tuning, unlike LLaMA2-13B, which improved in MCQ explanation generation through instruction fine-tuning.

\subsection{MCQ Explanation Generation} We employed instruction fine-tuning on LLaMA2-13B across all subjects to train an explanation generator, comparing it to four baseline models: Vicuna-13B, Vicuna-13B fine-tuned on each subject, LLaMA2-13B fine-tuned on each subject, and both GPT-3.5 and GPT-4. Given the cost of calling the GPT-4 API, we randomly selected 100 samples from the entire test set. We evaluated the syntactic and semantic similarity of generated explanations to ground truth explanations (student-authored) using BLEU~\citep{papineni2002bleu} and BERT scores~\citep{zhang2019bertscore}, respectively. In our experiments, GPT-4 consistently outperformed other models, achieving the highest BLEU and BERT scores across the majority of datasets, as shown in Table~\ref{explanation-generator-comparison}. Further investigation revealed that instruction fine-tuning significantly improved both BLEU and BERT scores for Vicuna-13B compared to the unmodified version. Extending this fine-tuning approach to LLaMA2-13B led to even more promising results. Specifically, instruction fine-tuned LLaMA2-13B surpassed Vicuna-13B and even outperformed GPT-4 in certain tasks. It achieved higher scores in the Sydney Biology and Auckland Law subjects and outperformed GPT-3.5 in four out of five datasets, with the exception of the UK Medical Year 2 subject. Two fine-tuning strategies were applied to LLaMA2-13B: individual fine-tuning for each task (``Fine-tuned LLaMA2-13B'') and a merged fine-tuning approach, combining training sets from all five tasks (``Fine-tuned LLaMA2-13B Merged''). The merged approach performed best on the Auckland Law subject, likely because it incorporated four biology/medicine datasets, increasing the diversity of topics and training data. Our findings suggest that both instruction fine-tuned Vicuna-13B and LLaMA2-13B effectively learned to mimic the characteristics of student-generated explanations.

\subsection{MCQ Explanation Evaluation} 
Table~\ref{explanation-evaluator-comparison} presents a comparison of fine-tuned and non-fine-tuned LLaMA2-13B models and GPT-4 on 100 randomly selected test cases from Sydney and Cardiff Biology, Auckland Law, and UK Medical Year 1 and 2 subjects for the MCQ explanation evaluation task. Using question quality rating labels, we trained a model to evaluate explanations by replacing the MCQ explanations. Mean Squared Error (MSE) was used as the metric, with a lower score indicating closer alignment to the ground truth. As shown, both fine-tuned LLaMA2-13B models significantly outperform the baselines in terms of MSE, suggesting they better capture the distribution of student-generated ratings. The ``Fine-tuned LLaMA2-13B Merged'' model achieves better performance than the ``Fine-tuned LLaMA2-13B,'' indicating that incorporating diverse subject data enhances predictive accuracy. In contrast, LLaMA2-13B without task-specific fine-tuning and GPT-4 underperform, often inflating scores, likely due to biases introduced by Reinforcement Learning from Human Feedback (RLHF). These findings highlight the importance of instruction fine-tuning for improving model performance in educational feedback applications.

\begin{table}[h]
\centering
\resizebox{\columnwidth}{!}{
\Huge
\begin{tabular}{@{}lcccc@{}}
\toprule
\begin{tabular}[c]{@{}l@{}}Models $\rightarrow$ \\ Metrics $\downarrow$\end{tabular} & LLaMA2-13B & \begin{tabular}[c]{@{}c@{}}Fine-tuned\\ LLaMA2-13B\end{tabular} & \begin{tabular}[c]{@{}c@{}}Fine-tuned\\ LLaMA2-13B\\Merged \end{tabular} & GPT-4  \\ \midrule
\multicolumn{5}{c}{Sydney Biology Subject}                                                                                                        \\
MSE                                            &   1.21   &    0.43                                                  & \textbf{0.22}     & 3.95 \\ \midrule
\multicolumn{5}{c}{Cardiff Biology Subject}                                                                                                       \\
MSE                                            &  0.58   &     0.10                                              & \textbf{0.09}       & 3.28 \\ \midrule
\multicolumn{5}{c}{Auckland Law Subject}                                                                                                          \\
MSE                                            &  2.86    &     \textbf{0.11}
                                        &   0.12         & 0.42 \\\midrule
\multicolumn{5}{c}{UK Medical Year 1 Subject}                                                                                                      \\
MSE                                            &  0.84    &    0.19                                       &  \textbf{0.15}            & 3.23 \\\midrule
\multicolumn{5}{c}{UK Medical Year 2 Subject}                                                                                                   \\
MSE                                            &  1.71
    &   0.10                                      &  \textbf{0.09}             & 3.02 \\\bottomrule
\end{tabular}}
\caption{We compared the fine-tuned LLaMA2-13B with the non-fine-tuned LLaMA2-13B and GPT-4 on 100 test cases for MCQ explanation evaluation.}
\label{explanation-evaluator-comparison}
\end{table}
\section{Conclusions and Future Work}

This study introduces the "ILearner-LLM" framework, which uses large language models to generate and assess explanations for learner-sourced multiple-choice questions. Experiments show that our iterative enhancement approach improves explanation quality, with better BLEU and BERT scores for LLaMA2-13B and GPT-4 compared to fine-tuned versions. To increase explanation diversity, we will explore different temperature hyperparameters in "ILearner-LLM."

\bibliography{aaai25}

\end{document}